%% file: main.tex
\documentclass[runningheads]{llncs}
\usepackage[T1]{fontenc}
\usepackage{graphicx}
\usepackage{bm}
\usepackage{mdframed}
\usepackage{multirow}
\usepackage{subcaption} 
\usepackage{soul}
\usepackage{rotating}
\usepackage{hyperref}
\usepackage{amsmath}
\usepackage{wrapfig}
\usepackage{booktabs}
\usepackage{amsfonts}
\usepackage{orcidlink}  
\usepackage[misc,geometry]{ifsym}

\usepackage[most]{tcolorbox}{\normalsize}
\usepackage[ruled,linesnumbered]{algorithm2e} 
\usepackage{color}

\begin{document}
\title{Graph Retrieval-Augmented LLM for Conversational Recommendation Systems}
\titlerunning{Graph Retrieval-Augmented LLM for CRS}

\author{
  Zhangchi Qiu\inst{1}{\orcidlink{0000-0001-8763-4070}} \and
  Linhao Luo\inst{2}\orcidlink{0000-0003-0027-942X} \and
  Zicheng Zhao\inst{3}\orcidlink{0000-0002-5195-6542}
  \and
  Shirui Pan\inst{1}\orcidlink{0000-0003-0794-527X} \and 
  Alan Wee-Chung Liew\inst{1}\textsuperscript{(\Letter)}\orcidlink{0000-0001-6718-7584}
}

\authorrunning{Z. Qiu et al.}
%
\institute{
  Griffith University, Gold Coast, Australia\\
  \email{zhangchi.qiu@griffithuni.edu.au}, \\ \email{\{s.pan, a.liew\}@griffith.edu.au} \and
  Monash University, Melbourne, Australia\\
  \email{linhao.luo@monash.edu} \and
  Nanjing University of Science and Technology, Nanjing, China\\
  \email{zicheng.zhao@njust.edu.cn}
}

\maketitle              
\begin{abstract}
Conversational Recommender Systems (CRSs) have emerged as a transformative paradigm for offering personalized recommendations through natural language dialogue. However, they face challenges with knowledge sparsity, as users often provide brief, incomplete preference statements. While recent methods have integrated external knowledge sources to mitigate this, they still struggle with semantic understanding and complex preference reasoning. Recent Large Language Models (LLMs) demonstrate promising capabilities in natural language understanding and reasoning, showing significant potential for CRSs. Nevertheless, due to the lack of domain knowledge, existing LLM-based CRSs either produce hallucinated recommendations or demand expensive domain-specific training, which largely limits their applicability. In this work, we present \textbf{G-CRS} (\textbf{G}raph Retrieval-Augmented Large Language Model for \textbf{C}onversational \textbf{R}ecommender \textbf{S}ystem), a novel training-free framework that combines graph retrieval-augmented generation and in-context learning to enhance LLMs' recommendation capabilities.
Specifically, G-CRS employs a two-stage retrieve-and-recommend architecture, where a GNN-based graph reasoner first identifies candidate items, followed by Personalized PageRank exploration to jointly discover potential items and similar user interactions. These retrieved contexts are then transformed into structured prompts for LLM reasoning, enabling contextually grounded recommendations without task-specific training.
Extensive experiments on two public datasets show that G-CRS achieves superior recommendation performance compared to existing methods without requiring task-specific training.

\end{abstract}
\input{section/intro.tex}
\input{section/pre.tex}
\input{section/method.tex}
\input{section/results.tex}
\input{section/related_work.tex}
\input{section/conclusion}

\bibliographystyle{splncs04}
\bibliography{ref}
\end{document}

%% file: section/intro.tex
\section{Introduction}
Conversational Recommender Systems (CRSs) have emerged as a transformative paradigm that engages users in natural language dialogue to understand user preferences and provide personalized recommendations~\cite{jannach_survey_2021}. 
However, a key challenge in CRSs lies in the limited expression of user preferences within conversations, where users usually express their needs through brief and incomplete statements~\cite{zhou_improving_2020}.
This user-driven limitation leads to knowledge sparsity in CRSs, making it difficult for the system to fully comprehend user needs.
To address this knowledge sparsity, recent works have explored incorporating external knowledge sources~\cite{chen_towards_2019,zhou_improving_2020,qiu_knowledge_graphs_2024} to supplement the limited dialogue-level information.
While these knowledge-enhanced methods enhance CRS by providing valuable domain information, they still struggle to comprehend semantic nuances in dialogues and perform complex reasoning about user preferences, resulting in a superficial understanding of user needs.
The emergence of Large Language Models (LLMs) presents a promising direction for addressing the semantic understanding limitations, thanks to their remarkable capabilities in natural language understanding and complex reasoning~\cite{zhao_survey_large_language_models_2023,dubey_llama3herdmodels_2024,zheng_large_2025}. 
Recent works have explored leveraging LLMs for dialogue understanding and response generation in CRS tasks~\cite{feng_largelanguagemodelenhanced_2023}.
Despite the success, LLMs still fall short in delivering effective recommendations due to the lack of domain-specific knowledge~\cite{he_large_language_2023}. Trained on general data, LLMs struggle with conducting domain-specific recommendations, often producing hallucinated items or failing to capture user preferences due to insufficient understanding of item relationships and collaborative patterns~\cite{sun_llm_cf_2024}. One approach harnesses graph structure to reduce hallucinations~\cite{luo2024graph}.
Other approaches enable LLMs to acquire domain-specific knowledge and perform recommendation tasks through specialized training procedures~\cite{Yang_unleasing_2024,qiu2024unveilinguserpreferencesknowledge}. However, the training of LLMs demands substantial computational resources, which limits their applicability.

Recent advances in retrieval-augmented generation (RAG)~\cite{gao2024retrievalaugmentedgenerationlargelanguage} and in-context learning (ICL)~\cite{dong2024surveyincontextlearning} have shown promising results in enhancing the performance of LLMs in various tasks without training. RAG retrieves domain knowledge, while ICL offers few-shot task demonstrations, presenting a cost-effective way to enhance LLM recommendation capabilities. However, unlike traditional document retrieval tasks where semantic similarity is enough, applying RAG to CRS requires retrieving similar dialogues that share similar user preferences across multiple conversation turns~\cite{dao_broadening_2024}.

In this paper, we introduce \textbf{G-CRS} (\underline{\textbf{G}}raph Retrieval-Augmented Large Language Model for \underline{\textbf{C}}onversational \underline{\textbf{R}}ecommendation \underline{\textbf{S}}ystems), a novel training-free framework, 
that leverages both graph-enhanced RAG and ICL to effectively retrieve information for recommendations and eliminate the need for extensive model training.
G-CRS employs a two-stage retrieve-and-recommend framework that enhances traditional RAG approaches through graph structure to better capture item relationships and collaborative patterns for recommendation. Specifically, a GNN-based graph reasoner is adopted to capture the latent recommendation patterns from graphs and identify an initial set of candidate items based on entities mentioned in the conversation. These retrieved candidates, together with the mentioned entities, then serve as seed nodes for the Personalized PageRank (PPR) algorithm~\cite{Haveliwala_ppr_2002}, to further explore the graph structure and discover both potential-interest items and history conversations exhibiting similar user interest in a single retrieval step. The retrieved history conversations are used as the few-shot demonstration to guide LLMs to capture the preference in the current dialogue with ICL. This graph-based exploration significantly improves upon traditional RAG approaches by capturing both semantic relationships between items and collaborative patterns across user interactions. The retrieved conversations and item candidates are then transformed into structured prompts for LLM reasoning, enabling the model to leverage its powerful ICL capabilities for effective recommendation without any task-specific training. 

Our main contributions can be summarized as follows. First, we present G-CRS, a novel graph retrieval-augmented framework that leverages ICL to enhance LLMs for conversational recommendation without task-specific training. Second, we introduce a graph-based retrieval mechanism that unifies item discovery and examples retrieval through graph exploration, enabling LLMs to perform contextually grounded recommendations. Finally, extensive experiments on two public datasets show that our framework achieves superior recommendation accuracy compared to existing methods without requiring additional LLM training.

%% file: section/pre.tex
\section{Preliminaries}
\noindent\textbf{Conversational Recommendation}.
A CRS enables an interactive dialogue between a user and the system, formally denoted as a sequence $\mathcal{C} = [c_1, ..., c_T]$, where each message $c_t$ denotes either a user query or system response.
At dialogue turn $t$, we define the conversation history as $H_t = [c_1\text{:}c_t]$. The system's objective is to generate recommendations $\mathcal{I}_{t+1} \subseteq \mathcal{I}$ based on $H_t$, where $\mathcal{I}$ represents the complete item set. These recommendations inform the system's response $c_{t+1}$. When recommendations are not needed, $\mathcal{I}_{t+1}$ may be empty and the system focuses on preference elicitation or maintaining dialogue context.

\noindent\textbf{Graph Retrieval-Augmented Recommendation}.
While traditional RAG enhances LLMs through document retrieval, it primarily relies on semantic similarity, which is insufficient for conversational recommendation where items are interconnected through complex relationships and collaborative patterns.
To capture these crucial structural relationships, we formalize the domain knowledge as a knowledge graph $\mathcal{G} = (\mathcal{E}, \mathcal{R}, \mathcal{A})$, where the entity set $\mathcal{E}$ encompasses both items $\mathcal{I}$ and their attributes (e.g., actors, genres), with $\mathcal{I} \subseteq \mathcal{E}$. The relation set $\mathcal{R}$ defines connection types between entities, while $\mathcal{A}$ captures their relationships through an adjacency matrix.
For a conversation history $H_t$, let $\mathcal{E}_t \subseteq \mathcal{E}$ denote mentioned entities. 
We can construct a conversation-entity interaction graph that connects the history conversation corpus to their mentioned entities and successful recommendations.
By exploring these structural connections from mentioned entities, we retrieve both relevant item candidates and similar conversations:
\begin{equation}
\mathcal{I}_k, \mathcal{C}_n = \text{G-retriever}(H_t, \mathcal{G}),
\end{equation}
where $\mathcal{I}_k \subseteq \mathcal{I}$ represents the top $k$ retrieved candidate items, and $\mathcal{C}_n$ denotes the top $n$ similar conversation interactions from the history conversation corpus $\mathcal{M}=\{\mathcal{C}_1,\ldots,\mathcal{C}_{N}\}$.
These retrieved contexts enable the LLM to generate recommendations with in-context learning:
\begin{equation}
\mathcal{I}_{t+1} = \text{LLM}(H_t, \mathcal{I}_k, \mathcal{C}_n),
\end{equation}
where $\mathcal{I}_{t+1} \subseteq \mathcal{I}_k$ represents the final recommendations.

\begin{figure}[t!]
  \centering
  \includegraphics[width=1\linewidth]{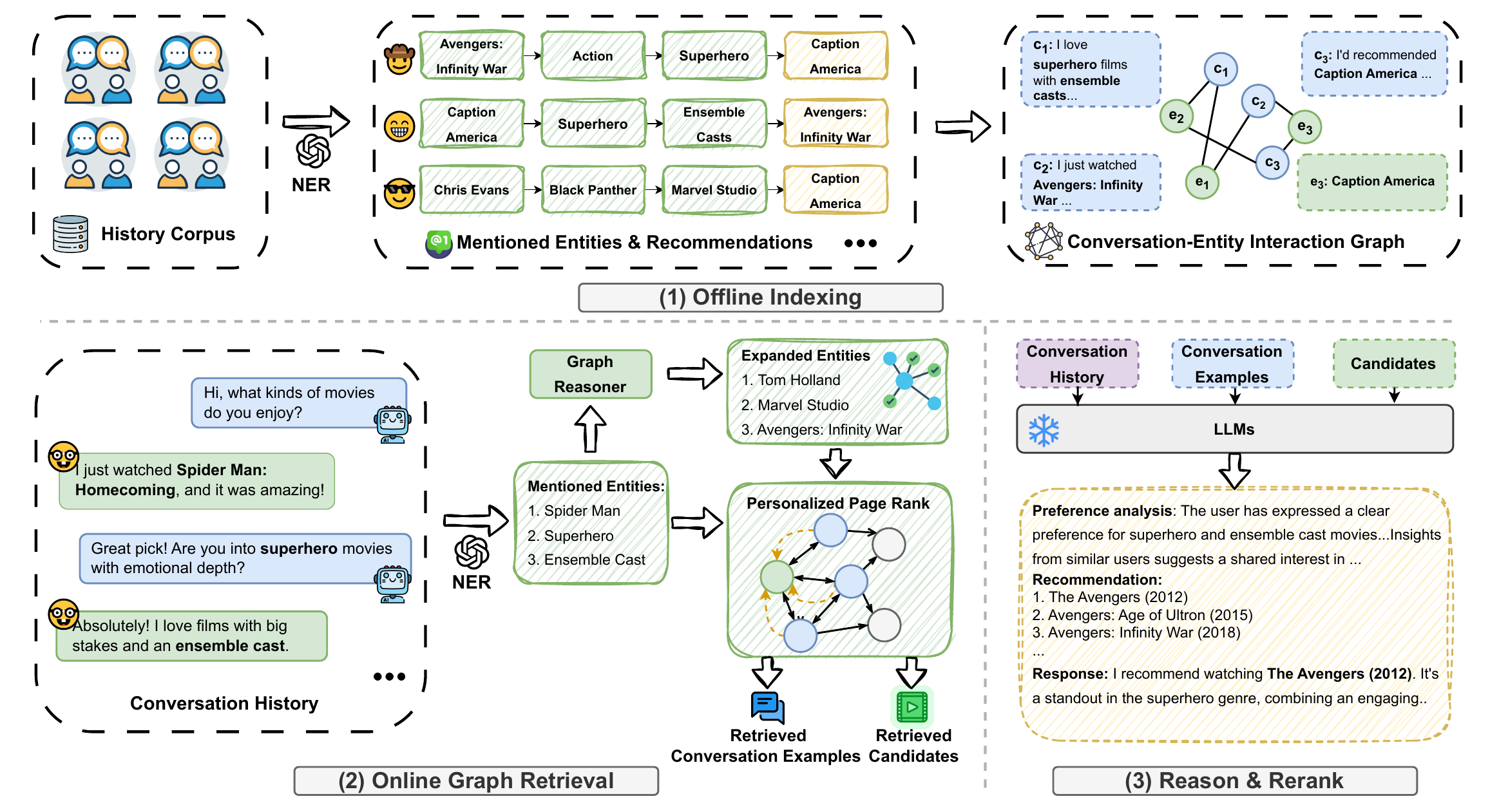}
  \vspace{-0.5cm}
  \caption{The overall framework of our G-CRS, a training-free framework for conversational recommendation. Our approach operates in three stages: (1) Offline Indexing: building a conversation-entity interaction graph from the training corpus; (2) Online Graph Retrieval: using mentioned entities as seed nodes for graph-based joint retrieval of similar conversations and candidate items; and (3) Reason \& Rerank: leveraging LLMs to analyze retrieved context and generated recommendations.}
  \label{fig:framework}
\end{figure}

%% file: section/method.tex
\section{Methodology}

Figure~\ref{fig:framework} illustrates an overview of the proposed G-CRS framework. The process begins with offline indexing, where entities and recommendations from the training corpus are extracted to construct a conversation-entity interaction graph. 
During online inference, given a conversation history, the system extracts mentioned entities and employs a graph reasoner to identify relevant entities, which together serve as seed nodes for Personalized PageRank (PPR) exploration. This graph retrieval process simultaneously discovers relevant conversation examples and candidate items. The retrieved information is then fed to the LLM for preference analysis and recommendation generation.

\subsection{Offline Indexing}

The offline indexing phase employs a graph-structured representation to capture both item relationships and conversational patterns, enabling efficient retrieval of contextually relevant recommendations and example conversations. 
Unlike traditional retrieval approaches that rely solely on semantic similarity, conversational recommendation requires capturing complex user preference patterns expressed across multiple dialogue turns, where both items and user interactions need to be effectively represented.
Specifically, we construct a Conversation-Entity Interaction Graph through entity linking and graph construction. 
First, a Named Entity Recognition (NER) module processes the training dialogue corpus to identify and link mentioned entities to a predefined knowledge graph $\mathcal{G}$. 
For each conversation in the training corpus, the module extracts both mentioned entities and successful recommendations, establishing connections between dialogue contexts and items.
Through this entity linking process, we construct a frequency matrix $\mathbb{P} \in \mathbb{R}^{|\mathcal{E}| \times |\mathcal{M}|}$, where $|\mathcal{M}|$ is the total number of conversation histories across all history conversations corpus and each entry $p_{ij}$ indicates the number of times entity $e_i \in \mathcal{E}$ appears in conversation history $H_j$. This matrix captures the density of entity mentions across conversations, enabling retrieval based on entity-level relevance during inference.
Based on these entity-linked conversations, the Conversation-Entity Interaction Graph is constructed to capture three critical relationships: \emph{entity-conversation mentions}, \emph{entity-entity co-occurrences}, and \emph{conversation-recommendation links}. This unified graph structure serves as the foundation for online graph retrieval by preserving both recommendation success patterns and the contextual flow of how entities are discussed in conversations.

\subsection{Online Graph Retrieval}

During online inference, G-CRS employs a multi-stage graph retrieval mechanism to identify both relevant candidate items and similar dialogue examples that can guide the LLM's recommendation reasoning. This process consists of entity linking, entity expansion through graph reasoning, and unified retrieval via PPR~\cite{Haveliwala_ppr_2002}. 
Given a conversation history $H_t$, we first employ an entity extractor to identify mentioned entities $\mathcal{E}_t$ in the dialogue, following the same approach used in our offline indexing phase. However, as users typically express their preferences through brief and incomplete statements~\cite{jannach_survey_2021,zhou_improving_2020}, relying solely on explicitly mentioned entities often leads to a limited understanding of user interests.
To address this preference sparsity, we leverage a pretrained graph reasoner~\cite{chen_towards_2019} to explore the knowledge graph structure to identify semantically related entities $\mathcal{E}'_t$ that align with potential user interests:
\begin{equation}
\mathcal{E}'_t = \text{G-Reasoner}(\mathcal{E}_t, \mathcal{G}),
\end{equation}
where the reasoner considers both direct relationships and higher-order connections in the knowledge graph to augment the initial preference signals. 
This expansion step is crucial for capturing implicit user preferences that may not be directly expressed in the conversation but are likely relevant based on domain knowledge encoded in the graph structure.
We denote the augmented entity set as:
\begin{equation}
\tilde{\mathcal{E}}_t = \mathcal{E}_t \cup \mathcal{E}'_t.
\end{equation}
This augmented entity set $\tilde{\mathcal{E}}_t$, then served as seed nodes for the PPR algorithm~\cite{Haveliwala_ppr_2002} to explore the Conversation-Entity Interaction Graph for retrieving both relevant items and similar conversations.
Specifically, the PPR algorithm computes a relevance score vector $\mathbf{r} \in \mathbb{R}^{|\mathcal{E}| \times 1}$ over all nodes in the graph, where each entry represents the importance of that node with respect to the seed nodes.
The initial personalization vector $\mathbf{p}$ is defined as:
\begin{equation}
p_i = \begin{cases} 1 & \text{if node } i \in \tilde{\mathcal{E}}_t, \\
0 & \text{otherwise}.
\end{cases}
\end{equation}
The PPR scores $\mathbf{r}$ are computed through the iterative equation:
\begin{equation}
\mathbf{r} = \alpha\mathbf{p} + (1-\alpha)\mathcal{A}'\mathbf{r},
\end{equation}
where $\alpha \in (0, 1)$ is the teleport (restart) probability, $\mathcal{A}'=\mathcal{A}\mathcal{D}^{-1}$ is the row-normalized adjacency matrix, and $\mathcal{D}$ is the degree of each node.
From the resulting score vector $\mathbf{r}$, we extract the top-$k$ candidate items and top-$n$ similar conversation histories based on their PPR scores:
{
\begin{gather}
\mathcal{I}_k = \mathop{\mathrm{arg\,top}\text{-}k} \mathbf{r_{1:|\mathcal{I}|}}, \\
\mathcal{C}_n = \mathop{\mathrm{arg\,top}\text{-}n} \mathbf{r}^\top\mathbb{P},
\end{gather}
where $\mathbb{P}$ is the pre-computed entity-conversation frequency matrix. 
The score $\mathbf{r_{1:|\mathcal{I}|}}$ represents the relevance of items, and $\mathbf{r}^\top\mathbb{P}$ aggregates the relevance scores over conversations.
Based on these scores, we retrieve top-$k$ items $\mathcal{I}_k$ and top-$n$ conversations $\mathcal{C}_n$ to support LLM reasoning.
}

\subsection{Retrieval-Augmented Reasoning and Recommendation}
After retrieving candidate items $\mathcal{I}_k$ and similar user interactions $\mathcal{C}_n$ through graph-structured exploration, we leverage the LLM's reasoning capabilities to perform re-ranking with in-context learning.
Our approach transforms the retrieved information into structured prompts that enable the LLM to learn from successful recommendation patterns while maintaining computational efficiency.

\noindent\textbf{Structured Prompting for In-Context Learning.}
We construct structured prompts $P_t$ that combine conversation history $H_t$, retrieved interactions $\mathcal{C}_n$, and candidate items $\mathcal{I}_k$ to guide the LLM's reasoning process:
\begin{tcolorbox}[width=0.5\textwidth, center, colback=cyan!10!white, colframe=cyan!60!black]
\small
Prompt: $P_t$\\
Conversation history: $H_t$ \\
Conversation examples: $\mathcal{C}_n$\\
Item candidates: $\mathcal{I}_k$
\end{tcolorbox} 
\noindent The retrieved interactions $\mathcal{C}_n$ serve as demonstrations of successful recommendation patterns, showcasing how user preferences evolved across conversation turns and led to effective recommendations. The candidate items $\mathcal{I}_k$ provide a grounded space for item selection, ensuring the LLM's recommendations remain anchored to available items. This structured format enables the LLM to learn from similar interactions while constraining its outputs to valid items within the catalog.

\noindent\textbf{Context-Aware Reranking.}
The LLM performs contextual reasoning over the prompted information $P_t$ to analyze and rerank candidate items $\mathcal{I}_k$:
\begin{equation}
\mathcal{I}_{t+1}, R_t = \text{LLM}(P_t),
\end{equation}
where $\mathcal{I}_{t+1} \subseteq \mathcal{I}_k$ represents the ranked recommendations and $R_t$ denotes the corresponding reasoning explanation.

%% file: section/results.tex
\section{Experiments}

\subsection{Experimental Setup}
\noindent\textbf{Datasets.} We conduct our experiments on two widely used CRS datasets: ReDial~\cite{li_towards_deep_2018} and INSPIRED~\cite{hayati-etal-inspired-2020}. 
ReDial includes 11,348 dialogue sessions focused on movie recommendations, collected via crowd-sourced interactions on Amazon Mechanical Turk (AMT). 
INSPIRED consists of 999 movie recommendation dialogues and incorporates social science-based recommendation strategies.
We constructed the knowledge graph by scraping data from IMDB\footnote{https://www.imdb.com/}, using movie titles and their corresponding release years as the primary identifiers.

\noindent\textbf{Evaluation Metrics.}
Following previous work~\cite{zhou_improving_2020,wang_towards_2022}, we use Hit Ratio (HR@$K$) and Mean Reciprocal Rank (MRR@$K$) with $K$=10 and 50 as metrics. HR@$K$ measures if ground truth items appear in the top-$K$ recommendations, while MRR@$K$ evaluates their average reciprocal rank positions.

\noindent\textbf{Baseline Methods.}
We evaluate our approach against three categories of baselines.
The first category comprises zero-shot retrieval methods: \emph{BM25}\cite{robertson_2009} and \emph{Sentence-BERT}\cite{reimers-2019-sentence-bert}.
The second category includes language models fine-tuned for recommendations, including \emph{BERT}~\cite{devlin_bert_2019}, \emph{GPT-2}~\cite{radford2019language}, and \emph{Llama3.1-8B}~\cite{dubey_llama3herdmodels_2024}.
The third category comprises specialized CRS models: 
\emph{ReDial}~\cite{li_towards_deep_2018}, 
\emph{KBRD}~\cite{chen_towards_2019}, 
\emph{KGSF}~\cite{zhou_improving_2020},
\emph{BARCOR}~\cite{wang2022barcorunifiedframeworkconversational}, 
\emph{UniCRS}~\cite{wang_towards_2022},
\emph{COLA}~\cite{lin_cola_2023}, and
\emph{PECRS}~\cite{ravaut-etal-2024-parameter}.

\noindent\textbf{Implementation Details.}
For baseline models, we adopt most implementations from either CRS-Lab~\cite{zhou-etal-2021-crslab} or authors' publicly released code repositories to ensure a fair comparison. 
For COLA~\cite{lin_cola_2023}, since its source code is not publicly available, we report the paper's official performance.
Our framework uses GPT-3.5 to extract entities from conversations and a pretrained graph reasoner~\cite{chen_towards_2019} for initial candidate retrieval, followed by PPR expansion to retrieve top-100 and top-150 candidates for LLM reranking.
Three similar conversation examples are retrieved for ICL.
We employ GPT-4o\footnote{gpt-4o-2024-08-06} for reasoning and recommendation generation and fuzzy string\footnote{https://github.com/seatgeek/thefuzz} matching for title matching during evaluation. 
\subsection{Recommendation Evaluation}
Table~\ref{tab:rec_results} presents the comprehensive evaluation comparing G-CRS against various baseline methods. 
Zero-shot retrieval methods (BM25 and Sentence-BERT) show limited performance as they rely solely on lexical or semantic matching without understanding user preferences.
Fine-tuned language models achieve better results through supervised training, with Llama3.1-8B showing strong performance (HR@10 = 0.188 on ReDial) but still falling short of specialized CRS approaches due to limited knowledge about items.
Through incorporating external knowledge, models like KBRD and KGSF achieve better results by enhancing entity representation. 
UniCRS further advances performance through unified prompt learning, showing strong results particularly on INSPIRED (HR@10 = 0.250, HR@50 = 0.408). 
The effectiveness of retrieval-enhanced approaches is exemplified by COLA, which achieves competitive performance on ReDial (HR@10 = 0.221) by leveraging relevant entities and items from similar conversations.
Our proposed G-CRS framework achieves consistent improvements over all baseline methods without any task-specific training, demonstrating that our graph-structured retrieval mechanism more effectively captures both item relationships and user preferences while enabling LLM to leverage these patterns through in-context learning.
\input{table/rec_table}

\subsection{Ablation Studies}
\input{table/ab_rag}

To evaluate the effectiveness of different components in G-CRS, we conduct ablation experiments by removing or replacing key components, as shown in Table~\ref{tab:ablation}. Removing the graph reasoner (w/o G-reasoner) leads to substantial performance degradation across all metrics, with HR@10 dropping by 18.0\% and 5.9\% on ReDial and INSPIRED respectively, demonstrating its crucial role in expanding the limited user-mentioned entities into a more comprehensive candidate set. The PPR component shows moderate impact, suggesting its value in discovering relevant items and conversations through graph exploration. Removing retrieved conversation examples (w/o ICL) leads to performance deterioration, particularly in ranking quality, demonstrating that example conversations help the LLM better understand recommendation patterns through in-context learning. 
When replacing our graph-enhanced retrieval with traditional methods, both BM25 and Sentence-BERT show significant performance drop (HR@10 decreasing by up to 27.9\% and 22.1\% respectively), highlighting the limitation of purely lexical or semantic retrieval methods compared to our graph-based method in capturing both structural relationships and collaborative patterns crucial for conversational recommendation.
\begin{figure}[t]
    \centering
    \begin{subfigure}[b]{0.240\textwidth}
        \centering
        \includegraphics[width=\textwidth]{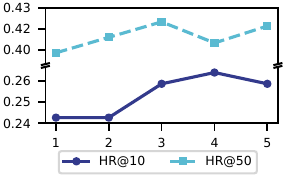}
        \caption{HR vs ICL}
        \label{fig:hr_icl}
    \end{subfigure}
    \begin{subfigure}[b]{0.240\textwidth}
        \centering
        \includegraphics[width=\textwidth]{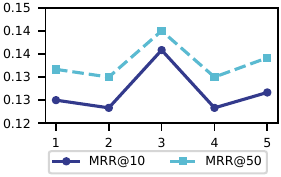}
        \caption{MRR vs ICL}
        \label{fig:mrr_icl}
    \end{subfigure}
    \begin{subfigure}[b]{0.240\textwidth}
        \centering
        \includegraphics[width=\textwidth]{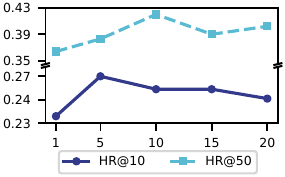}
        \caption{HR vs Entity}
        \label{fig:hr_node}
    \end{subfigure}
    \begin{subfigure}[b]{0.240\textwidth}
        \centering
        \includegraphics[width=\textwidth]{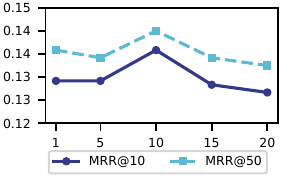}
        \caption{MRR vs Entity}
        \label{fig:mrr_node}
    \end{subfigure}
    \vspace{-0.2cm}
    \caption{Impact of varying (a)-(b) number of in-context learning examples and (c)-(d) number of retrieved entities from graph reasoner on INSPIRED dataset.}
    \label{fig:ablation_analysis}
    \vspace{-0.2cm}
\end{figure}

\noindent\textbf{Impact of ICL Examples.}
We further analyze how the number of in-context learning examples affects model performance, as shown in  Figures~\ref{fig:hr_icl} and~\ref{fig:mrr_icl}.
Our experiments reveal that both metrics improve with more examples up to a certain point, after which the performance plateaus or slightly decreases. This optimal range suggests that while retrieved examples are crucial for in-context learning, a moderate number of high-quality examples is sufficient for effective recommendation.

\noindent\textbf{Impact of Graph Expansion.}
The impact of entity set size retrieved from the graph reasoner on recommendation performance is illustrated in Figures~\ref{fig:hr_node} and~\ref{fig:mrr_node}. Our experiments demonstrate that both metrics improve significantly as the number of entities increases, with HR@50 rising from 0.36 to 0.42. However, further increasing the candidate set leads to performance plateauing or slight degradation. This suggests that while the graph reasoner's initial retrieval is crucial for seeding the subsequent PPR expansion, a moderate number of high-quality candidates provides the optimal foundation for effective recommendations.

\vspace{-0.2cm}
\subsection{Case Study}

The example above demonstrates G-CRS's capabilities in a real conversation scenario. 
Given a straightforward request for Mel Gibson movies, the system first leverages its graph retriever to identify relevant candidates from Gibson's filmography while simultaneously retrieving similar actor-based recommendation patterns.
Through in-context learning, G-CRS processes this single preference signal to generate ranked recommendations, prioritizing "Braveheart" while providing contextual details about Gibson's dual role as director and star. 
The generated response shows how G-CRS effectively combines graph-structured knowledge with conversation patterns to deliver informed recommendations even with minimal user input.

\begin{tcolorbox}[colback=green!5,colframe=green!40!black,fontupper=\scriptsize, title=Input] 
\textbf{Instructions}: <Instructions>

\textbf{Conversation History:} User: Good morning! I'm in the mood for a movie with Mel Gibson. Any suggestions

\textbf{Retrieved Example:} 

User: Film Noir, movies with Humphrey Bogart...

Recommender: Have you seen El Dorado (1966) with John Wayne...

[Actor-based recommendation pattern]

\textbf{Retrieved Candidates:}

The Patriot (2000), Mad Max (1979), Lethal Weapon (1987), ..., Braveheart (1995), ...
\end{tcolorbox}
\begin{tcolorbox}[colback=yellow!5,colframe=yellow!40!black,fontupper=\scriptsize, title=Output]
\textbf{Preference Analysis:} User explicitly expressed preference for Mel Gibson films. There are no additional explicit or implicit preferences provided in the conversation. Retrieved example shows a successful pattern of recommending diverse films featuring the requested actor...

\textbf{Top Reranked Recommendations:} 
1. Braveheart (1995), 2. The Patriot (2000), 3. Mad Max Beyond Thunderdome (1985) ...

\textbf{Generated Response:} I'd highly recommend `Braveheart' (1995), which is one of Mel Gibson's most iconic roles. He both directed and starred in this historical epic.
\end{tcolorbox}

%% file: table/rec_table.tex
\begin{table}[t!]
\centering
\caption{Performance comparison on recommendation tasks. The best results are highlighted in bold, and the second best results marked with an underline.}
\label{tab:rec_results}
\resizebox{0.8\linewidth}{!}{%
\begin{tabular}{@{}lcccccccc@{}}
\toprule
\multirow{2}{*}{Model} & \multicolumn{4}{c}{ReDial} & \multicolumn{4}{c}{INSPIRED} \\
\cmidrule(lr){2-5} \cmidrule(lr){6-9}
 & HR@10 & HR@50 & MRR@10 & MRR@50 & HR@10 & HR@50 & MRR@10 & MRR@50 \\
\midrule

  BM25 & 0.022 & 0.056 & 0.008 & 0.009 & 0.032 & 0.110 & 0.010 & 0.014 \\
  Sentence-BERT &  0.043 & 0.100 & 0.020 & 0.023 & 0.090 & 0.197 & 0.035 & 0.040 \\
  \midrule
  BERT & 0.143 & 0.319 & 0.052 & 0.059 & 0.179 & 0.328 & 0.072 & 0.079 \\
  GPT-2 & 0.147 & 0.327 & 0.051 & 0.056 & 0.112 & 0.278 & 0.063 & 0.076 \\
  Llama3.1-8B & 0.188 & 0.376 & 0.078 & 0.087 & 0.190 & 0.332 & 0.094 & 0.102\\
  \midrule
  ReDial & 0.140 & 0.320 & 0.035 & 0.045 & 0.117 & 0.285 & 0.022 & 0.048 \\
  KBRD & 0.151 & 0.336 & 0.071 & 0.079 &  0.172 & 0.265 & 0.086 & 0.091 \\
  KGSF & 0.183 & 0.378 & 0.072 & 0.081 & 0.175 & 0.273 & 0.088 & 0.093 \\
  BARCOR  & 0.169 & 0.374 & 0.063 & 0.073 & 0.185 & 0.339 & 0.080 & 0.087 \\
  UniCRS & 0.216 & 0.416 & \underline{0.087} & 0.095 & \underline{0.250} & \underline{0.408} & \underline{0.109} & \underline{0.117} \\
  COLA & \underline{0.221} & \underline{0.426} & 0.086 & \underline{0.096} & - & - & - & - \\
  PECRS & 0.205 & 0.399 & 0.083 & 0.093 & 0.179 & 0.337 & 0.084 & 0.092 \\
\midrule
 G-CRS & \textbf{0.244} & \textbf{0.426} & \textbf{0.099} & \textbf{0.108} & \textbf{0.254} & \textbf{0.420} & \textbf{0.139} & \textbf{0.144} \\
\bottomrule
\end{tabular}
}
\vspace{-0.5cm}
\end{table}

%% file: table/ab_rag.tex
\begin{table*}[t!]
\centering
\caption{Ablation study on model components. ``w/o'' indicates component removal, ``w/'' denotes replacement of graph-enhanced retrieval with alternative methods while retaining LLM reranking.}
\label{tab:ablation}
\resizebox{.8\linewidth}{!}{%
\begin{tabular}{@{}lcccccccc@{}}
\toprule
\multirow{2}{*}{Method} & \multicolumn{4}{c}{ReDial} & \multicolumn{4}{c}{INSPIRED} \\
\cmidrule(lr){2-5} \cmidrule(lr){6-9}
 & HR@10 & HR@50 & MRR@10 & MRR@50 & HR@10 & HR@50 & MRR@10 & MRR@50 \\
\midrule
G-CRS & \textbf{0.244} & \textbf{0.426} & \textbf{0.099} & \textbf{0.108} & \textbf{0.254} & \textbf{0.420} & \textbf{0.139} & \textbf{0.144} \\
\midrule
w/o G-Reasoner & 0.200 & 0.324 & 0.089 & 0.093 &  0.239 & 0.356 & 0.132 & 0.139 \\
w/o PPR & 0.239 & 0.412 & 0.090 & 0.099 &  0.254 & 0.384 & 0.134 & 0.140 \\
w/o ICL & 0.232 &  0.423 &  0.086 & 0.094 &  0.248 & 0.402 & 0.127 & 0.135 \\
\midrule
w/ BM25 &  0.176 & 0.286 & 0.075 & 0.078 & 0.182 & 0.265 & 0.115 & 0.119   \\
w/ Sentence-BERT &  0.190 & 0.314 & 0.080 & 0.090 & 0.205 & 0.318 & 0.108 & 0.115 \\
\bottomrule
\end{tabular}
}
\label{tab:ablation}
\end{table*}

%% file: section/related_work.tex
\section{Related Work}
\noindent\textbf{Conversational Recommendation.}
Conversational Recommender Systems (CRSs) deliver personalized recommendations by engaging users in natural language dialogues to understand their preferences~\cite{jannach_survey_2021}.
Early approaches leveraged knowledge graphs and reviews\cite{chen_towards_2019,zhou_improving_2020} to enhance recommendation quality, while pre-trained language models (PLMs)~\cite{wang_towards_2022,wang2022barcorunifiedframeworkconversational,ravaut-etal-2024-parameter} are employed to improve dialogue understanding and response generation. 
Recent retrieval-based CRSs like COLA~\cite{lin_cola_2023} and DCRS~\cite{dao_broadening_2024} leverage similar conversations to augment user preferences via entity-based matching or dense embeddings. However, this entity-centric augmentation, while providing additional signals, fails to capture the complete user experience context.
Building on this, LLMs enhance CRSs by enabling improved dialogue planning and response generation~\cite{feng_largelanguagemodelenhanced_2023}.
Although~\cite{he_large_language_2023} explores LLMs' zero-shot recommendation capabilities, grounding generated recommendations in real item spaces remains challenging. 
Different from previous LLM-enhanced CRS work, our framework leverages graph-enhanced retrieval and in-context learning to enable training-free, grounded recommendations through retrieved examples and candidates.

\noindent\textbf{Retrieval-Augmentation Generation and Recommendation.}
Retrieval-Augmented Generation (RAG) has emerged as a powerful paradigm for enhancing LLMs' capabilities by grounding their responses in retrieved information~\cite{gao2024retrievalaugmentedgenerationlargelanguage}. Unlike traditional RAG approaches that aim to retrieve relevant documents and facts to ground LLM responses, applying RAG to recommendation systems presents unique challenges as it requires capturing complex user preferences and item relationships~\cite{wu2024coralcollaborativeretrievalaugmentedlarge}. This challenge is further amplified in CRS, where preferences are not only expressed across multiple dialogue turns but also often incomplete and implicit. While~\cite{Yang_unleasing_2024} explored RAG-enhanced CRS through specialized fine-tuning methods, they require extensive model training. To our knowledge, we are the first to propose a training-free Graph RAG framework for CRSs, which leverages graph structures to jointly capture item relationships and user interaction patterns through unified graph-based retrieval.

%% file: section/conclusion.tex
\section{Conclusion}

In this paper, we introduce G-CRS, a novel training-free framework that leverages graph-based retrieval and ICL with LLMs for conversational recommendation.
To address knowledge sparsity in user preferences, we propose a unified graph-based retrieval mechanism that leverages PPR to jointly discover relevant items and similar conversations.
Through ICL with retrieved examples, G-CRS enables LLMs to perform recommendations without expensive training procedures. 
Experiments on two public datasets demonstrate superior performance over existing approaches. 
Future work includes extending the framework to multi-modal scenarios where visual and textual information can be jointly leveraged for more comprehensive recommendation experiences.